\pdfoutput=1

\documentclass[11pt]{article}

\usepackage[final]{acl}
\usepackage{tcolorbox}
\usepackage{enumitem}
\usepackage{lipsum}
\usepackage[svgnames]{xcolor}  

\usepackage{times}
\usepackage{latexsym}
\usepackage{multirow}
\usepackage[T1]{fontenc}
\usepackage[utf8]{inputenc}
\usepackage{microtype}
\usepackage{inconsolata}
\usepackage{graphicx}

\usepackage[table]{xcolor}
\usepackage{enumitem}
\usepackage{lipsum}
\usepackage{amsmath}
\usepackage{adjustbox}
\usepackage{arydshln} 

\definecolor{lightgreen}{RGB}{220,255,220}
\definecolor{lightorange}{RGB}{255,235,205}
\definecolor{lightblue}{RGB}{220,240,255}
\definecolor{mylightorange}{rgb}{1.0, 0.9, 0.8}

\usepackage{tabularx}
\usepackage{booktabs} 

\title{A Multi-Level Benchmark for Causal Language Understanding in Social Media Discourse}

\author{
  Xiaohan Ding \quad
  Kaike Ping \quad
  Buse Çarık \quad
  Eugenia Rho \\
  Department of Computer Science, Virginia Tech, USA \\
  \texttt{\{xiaohan,kaike,buse,eugenia\}@vt.edu}
}

\begin{document}
\maketitle
\begin{abstract}
Understanding causal language in informal discourse is a core yet underexplored challenge in NLP. Existing datasets largely focus on explicit causality in structured text, providing limited support for detecting implicit causal expressions, particularly those found in informal, user-generated social media posts.
We introduce \textbf{CausalTalk}, a multi-level dataset of five years of Reddit posts (2020–2024) discussing public health related to the COVID-19 pandemic, among which $10,120$ posts are annotated across four causal tasks: (1) binary causal classification, (2) explicit vs. implicit causality, (3) cause–effect span extraction, and (4) causal gist generation. Annotations comprise both gold-standard labels created by domain experts and silver-standard labels generated by GPT-4o and verified by human annotators.
CausalTalk bridges fine-grained causal detection and gist-based reasoning over informal text. It enables benchmarking across both discriminative and generative models, and provides a rich resource for studying causal reasoning in social media contexts\footnote{The dataset and code are available at \url{https://github.com/xding2/CausalTalk}}. 
\end{abstract}

\section{Introduction}
\begin{table*}[ht]
\centering

\begin{adjustbox}{width=0.9\textwidth}
\begin{tabular}{@{}lllll@{}}
\toprule
\textbf{Dataset} & \textbf{Language} & \textbf{\# Entries} & \textbf{\# Causality Relations} & \textbf{Causality Type} \\
\midrule
SemEval-2010 Task 8 \cite{hendrickx2009semeval} & English & 10,674 & 1,325 & Explicit \\
CausalTimeBank \cite{mirza2014annotating} & English & 11,000 & 318 & Explicit \\
EventStoryLine \cite{caselli2017eventstoryline} & English & 7,275 & 5,519 & Explicit, Implicit \\
BioCausal \cite{mihuailua2013biocause} & English & 13,342 & 7,562 & Explicit \\
BECauSE 2.0 \cite{dunietz2017because} & English & 1,803 & 1,634 & Explicit \\
RED \cite{o2016richer} & English & 8,731 & 4,969 & Explicit \\
CaTeRS \cite{mostafazadeh2016caters} & English & 2,708 & 488 & Explicit, Implicit \\
MAVEN-ERE \cite{wang2022maven} & English & 103,193 & 57,992 & Explicit \\
HPG Incidents \cite{inoue2023towards} & Japanese & 18,171 & 970 & Explicit \\
\bottomrule
\end{tabular}
\end{adjustbox}
\caption{Overview of major causal relation datasets used in NLP.}
\label{tab:causal_datasets}
\end{table*}
Causal language understanding is a foundational yet underexplored challenge in natural language processing (NLP)~\cite{niess2025addressing, yu2025causaleval, kim2023can, ali2021causality, blanco2008causal, heddaya2024causal}. Accurately identifying cause-effect relations, that is, how one event leads to another, is essential for tasks such as information extraction~\cite{dukic2023leveraging, hu2025large}, narrative understanding~\cite{sun2023event, zhang2021salience}, and decision support~\cite{ding2024leveraging}. While prior work has primarily focused on structured or formal texts~\cite{caselli2017eventstoryline, kyriakakis2019transfer}, causal reasoning in informal or unstructured social media text remains underdeveloped, despite its growing relevance for analyzing discourse around public health~\cite{son2018causal}, misinformation~\cite{adams2023misinformation}, and everyday decision-making~\cite{lin2008causal, ding2024leveraging}.

Existing causal datasets (Table~\ref{tab:causal_datasets}) have made progress by annotating explicit causal links across newswire~\cite{mirza2014annotating}, biomedical~\cite{mihuailua2013biocause}, and narrative texts~\cite{mostafazadeh2016caters}. However, they remain limited in several ways. (1) Most datasets emphasize explicit causality~\cite{hendrickx2009semeval, mirza2014annotating}, offering limited coverage of implicit relations that frequently appear in informal discourse~\cite{hartshorne2014implicit}. (2) Despite growing interest in long-context reasoning, existing benchmarks lack gist-driven causal annotations that reflect how humans distill causal meaning into concise representations~\cite{reyna2012new, brainerd1990gist}.

To address these gaps, we draw on fuzzy-trace theory (FTT)~\cite{reyna2012new, brainerd1990gist}, which argues that texts with coherent causal relationships help people form gist-like mental representations—concise abstractions that support better understanding and memory than verbatim information.
Recent work in LLM prompting has leveraged this insight to support long-context processing~\cite{lee2024human, dhaini2024explainability, liu2024improving}, but such approaches lack suitable training or evaluation data grounded in informal, real-world discourse. Inspired by these developments, we propose a new dataset to support both causal detection and gist-based generation.

We introduce \textbf{CausalTalk}, a multi-level dataset of Reddit posts from 2020 to 2024 that capture public health discussions, with a particular focus on the COVID-19 pandemic, annotated for causal language across four tasks: (1) binary causal classification, (2) explicit vs.\ implicit causality detection, (3) cause–effect span extraction, and (4) causal gist generation. The dataset includes both \textit{gold-standard} annotations—manually labeled and adjudicated by expert annotators—and \textit{silver-standard} annotations—generated by GPT-4o using zero-shot prompting and subsequently verified by human annotators, following established practices from prior annotation research~\cite{hengle2024still, li-etal-2024-maven, mirzakhmedova2024large}. In total, \textbf{CausalTalk} provides 10 million raw Reddit posts and 10,120 annotated instances across 43 public health related subreddits.

\section{Related Work}
\subsection{Causal Datasets in NLP}
Causal language understanding has long been a key challenge in information extraction. To support model development for causal relation identification, a range of datasets have been proposed across various domains and annotation schemes. A summary of these datasets is presented in Table~\ref{tab:causal_datasets}.

A foundational dataset often used for causal classification is the \textbf{SemEval-2010 Task 8} dataset \cite{hendrickx2009semeval}, which contains 10,674 sentences annotated with semantic relations between pairs of nominals, including 1,325 instances labeled as Cause–Effect. Although the dataset was originally developed for semantic relation classification, \cite{kyriakakis2019transfer} later adapted it for binary causal classification by labeling data with the Cause–Effect relation as causal and all others as non-causal.

The \textbf{CausalTimeBank} corpus \cite{mirza2014annotating} and the \textbf{EventStoryLine} dataset \cite{caselli2017eventstoryline} introduce event-level causal annotations by marking causal signals (C-SIGNAL) and links (CLINK), or PLOT LINKs (CAUSES, CAUSED BY). Following \cite{li2019knowledge}, \cite{kyriakakis2019transfer} filtered these datasets to include only intra-sentence causal pairs, resulting in 318 and 5,519 causal instances respectively. 

Other efforts have expanded the domain and annotation scope. For example, the \textbf{BioCausal} dataset provides biomedical causal annotations over 13,342 sentences from PubMed (7,562 causal), with a publicly available subset of 2,000 examples \cite{mihuailua2013biocause}.

To improve annotation granularity, the \textbf{BECauSE 2.0} corpus \cite{dunietz2017because} labels 1,803 instances of causal language across 5,380 sentences, with over 90\% of them including both cause and effect spans. The \textbf{Richer Event Description (RED)} corpus \cite{o2016richer} goes further to annotate 4,969 relations across 8,731 events in 95 documents, integrating causal, temporal, and coreference relations.

To explore both explicit and implicit causality, \textbf{CaTeRS} \cite{mostafazadeh2016caters} annotates 1,600 sentences from ROCStories, marking 488 causal and temporal relations across 2,708 events. Despite its limited size, it emphasizes narrative causality and implicit links, complementing the more structural datasets.

More recently, efforts have shifted toward large-scale and multilingual resources. \textbf{MAVEN-ERE} \cite{wang2022maven} is a unified event relation corpus with over 57,000 human-annotated causal links, 1.2M temporal relations, and 103,000 event coreference chains, offering substantial coverage and enabling joint modeling of event relations. Meanwhile, the \textbf{HPG Incidents} dataset \cite{inoue2023towards} focuses on industrial safety, with 970 annotated high-pressure gas incident reports designed for causal and named entity recognition tasks. Finally, the \textbf{MECI} corpus \cite{lai2022meci} advances event causality identification (ECI) research by offering consistent annotations across five typologically diverse languages (English, Danish, Spanish, Turkish, Urdu). With over 11,000 annotated relations, MECI supports multilingual and cross-lingual causality modeling. 
\subsection{Causal Gist and Fuzzy-Trace Theory in NLP and LLMs}

Fuzzy-trace theory (FTT) posits that people rely on simplified, essential meanings (called \textit{gists}), rather than exact details when processing information~\cite{brainerd1990gist}. This gist-based reasoning helps explain how individuals make decisions, assess risks~\cite{ding2024leveraging}, and comprehend long texts~\cite{lee2024human, reyna2012new}. Research shows that gist representations are more memorable than verbatim details~\cite{reyna2012new}, supporting the idea that cognitive development involves a shift from literal recall to abstract understanding~\cite{reyna2021scientific}.

In recent years, FTT has increasingly informed developments in natural language processing and large language models~\cite{ding2024leveraging, lee2024human, dhaini2024explainability, liu2024improving}. Studies show that summarizing long texts using gist-level representations, combined with attention to local details, improves the efficiency of long-context reasoning~\cite{lee2024human, reyna2012new}.
For example, \citeauthor{lee2024human} proposed \textit{ReadAgent}, a prompting-based LLM system designed to emulate human reading strategies. The system segments long texts into smaller episodes, generates gists for each segment, and maintains contextual links to support reasoning across extended content. A similar line of work,~\citeauthor{liu2024improving}~\citeyear{liu2024improving} uses summarization models to identify gists and integrates them into downstream models to enhance long-text understanding.

Extending these approaches, \citeauthor{ding2024leveraging}~\citeyear{ding2024leveraging} proposed a framework for applying causal gist generation to chain-of-thought prompting in LLMs. Their method, \textit{Role-Based Incremental Coaching} (RBIC), breaks down complex causal reasoning—especially within fragmented or informal online discourse—into more manageable subtasks, each designed to focus on a specific causal step and summarized by a causal gist. \citeauthor{ding2024leveraging}'s approach treats gist as a central step in structuring causal reasoning, allowing models to interpret nuanced online discourse.

Overall, past research highlights the importance of aligning causal language modeling with human cognitive strategies, especially gist-based understanding. Motivated by these insights, our dataset is designed to support multi-level causal understanding in social media discourse, ranging from detection to the generation of causal gist.

\section{Dataset Collection}

\subsection{Data Sources}
We collected Reddit posts (including submissions and their associated comments) related to public health, with a particular focus on the COVID-19 pandemic, from January 2020 through December 2024 across 43 subreddits. The primary data sources include pandemic-focused subreddits such as \texttt{r/Coronavirus}, \texttt{r/COVID19}, \texttt{r/wuhanflu}, and \texttt{r/DebateVaccines}. These 43 subreddits were selected due to their coverage of pandemic discussions from scientific, policy-related, and personal experience perspectives. Detailed descriptions of these subreddits are provided in the Appendix~\ref{app:A}.

We used the Pushshift API to retrieve data and filter it to ensure data quality. Specifically, duplicates, extremely short posts (fewer than 20 tokens), and non-English content were excluded. Finally, we anonymise the data user id to protect user privacy.

\begin{table}[!h]
\centering
\begin{adjustbox}{width=0.35\textwidth}
\begin{tabular}{lr}
\hline
\textbf{Feature} & \textbf{Count} \\
\hline
Total Submissions & 239,222 \\
Total Comments & 19,138,266 \\
Number of Subreddits & 43 \\
\hline
\end{tabular}
\end{adjustbox}
\caption{Summary Statistics of Reddit COVID-19 Pandemic Dataset}
\label{dataset_stats}
\end{table}

\subsection{Dataset Statistics}
The final dataset comprises 239,222 Reddit submissions and 19,138,266 comments collected over the five-year span. Table \ref{dataset_stats} summarizes the dataset statistics.

\section{Annotation Schema}

To capture the diverse manifestations of causal language in social media text, we designed a hierarchical annotation schema consisting of four complementary tasks: (1) binary annotation of causal language, (2) explicit versus implicit causality classification, (3) fine-grained span annotation of causes, effects, and signals, and (4) casual gist generation. Each task incrementally builds upon the previous, facilitating both analysis of causal structure and the creation of high-quality, multi-level benchmark data to support downstream NLP modeling.

\textbf{Task 1: Binary Annotation of Causal Language.} Our annotators received initial training based on \citeauthor{sanders2009introduction}~\citeyear{sanders2009introduction}'s definition of causal relationships in language, as well as the \textit{Annotation of Causal Relations} schema~\cite{rehbein-ruppenhofer-2017-catching}. Their research characterizes causal content as textual expressions indicating a relationship where one event, action, or condition directly leads to or influences another~\cite{rehbein-ruppenhofer-2017-catching, sanders2009introduction}. 

During the annotation process, each data entry was independently annotated by five individuals as either containing causal content (1) or not (0). Discrepancies were resolved through group discussions facilitated by a sixth annotator, with the final ground truth label determined by the majority vote reached during these discussions.

\textbf{Task 2: Explicit vs. Implicit Causality.} Our annotators were trained based on prior research to distinguish between \textit{Explicit} and \textit{Implicit} causality in language~\cite{sanders2009causal, pickering2007implicit}. The definitions and references used during training are as follows:

\begin{itemize}
    \item \textbf{Explicit causality}: Causality that is overtly expressed through causal markers such as \textit{“because,” “therefore,” “as a result,”} etc. These connective cues clearly indicate a causal relationship between two events or propositions~\cite{sanders2009causal}.
    \item \textbf{Implicit causality}: Causality that is not explicitly marked by connectors but is instead inferred from context, verb semantics, or the structure of events~\cite{pickering2007implicit}.
\end{itemize}

Each data entry that was labeled as containing causal content in Task 1 was further examined in Task 2. Specifically, these entries were annotated by five independent annotators as expressing either \textbf{explicit causality} (1) or \textbf{implicit causality} (0). As with the previous task, disagreements were resolved through group discussion led by a sixth annotator.

\textbf{Task 3: Cause–Effect Span Annotation.} In this task, annotators identified specific spans of text corresponding to the cause, effect, and the causal signal (when applicable) within each data entry labeled as containing causal content in Task 1 ($n = 600$). The goal of this span-level annotation was to capture the linguistic realization of causal relationships within individual sentences.

For instances of \textit{explicit causality}, annotators typically selected spans surrounding overt causal markers (e.g., “because,” “as a result”). In contrast, for \textit{implicit causality}, annotators relied on contextual cues, event structure, and verb semantics to infer and label the appropriate cause and effect spans, even when no explicit connective was present~\cite{rehbein-ruppenhofer-2017-catching}. Below are examples of annotated sentences demonstrating both explicit and implicit causal structures:

\textbf{Explicit causal signal:} \\
    \parbox{\columnwidth}{
        If <cause>\colorbox{lightgreen}{someone sneezes in your face} </cause> \\
        it can \colorbox{lightblue}{cause} 
        <effect> \colorbox{lightorange}{respiratory droplets that} \\
        \colorbox{lightorange}{ expose you to their virus-laden viruses} </effect>
    }
\\

  \textbf{Implicit causality (no signal):} \\
    \parbox{\columnwidth}{
        <cause>\colorbox{lightgreen}{Long-term isolation in the home}</cause>\\
        at least <effect>\colorbox{lightorange}{eliminates cross-contamination} </effect>\\
    }


As in previous tasks, all annotations were performed independently by five annotators. Discrepancies were discussed and resolved by a sixth annotator.

\textbf{Task 4: Gist Generation from Causal Relationships.} In the final task, annotators were asked to synthesize a \textbf{causal gist} that captures the core causal relationship(s) expressed in each sentence identified as containing causal content. 
The gist generation process was conducted in two stages. First, individual annotators independently drafted candidate gists based on the previously annotated \texttt{<cause>} and \texttt{<effect>} spans. Second, all annotators participated in a group discussion to review, refine, and consolidate these gists into a single agreed-upon version per instance. 


\subsection{Annotation Process}

We employed a dual annotation strategy to build both a high-quality gold-standard dataset and a large-scale silver-standard corpus.

\textbf{Gold Annotations.} Five annotators with backgrounds in causal linguistics and public health independently labeled 1,320 Reddit posts, randomly selected to ensure a balanced sample across 43 subreddits. Disagreements were resolved through group discussions led by a sixth annotator, resulting in consensus gold-standard labels for all annotation tasks. These tasks included causal classification ($600$ posts labeled as causal, $720$ as non-causal), explicit vs. implicit causality distinction ($243$ implicit, $357$ explicit), span extraction, and gist generation ($n = 600$ gists).

\textbf{Silver Annotations.} Building upon prior research in human–AI collaborative annotation~\cite{li-etal-2024-maven, mirzakhmedova2024large, chen2024large}, we utilized GPT-4o with zero-shot learning to generate silver-standard labels for $8,800$ Reddit posts (randomly selected to ensure a balanced sample across 43 subreddits). The model was prompted using gold-standard exemplars within an RBIC-based~\cite{ding2024leveraging} and chain-of-thought~\cite{wei2022chain} instruction format (see Appendix~\ref{app:B} for prompt details). All model-generated annotations were subsequently reviewed and refined by the same team of six expert annotators. Specifically, to assess the quality of the silver-standard annotations, the full set of $8{,}800$ posts was evaluated by human annotators according to the following dimensions:

\begin{itemize}
    \item \textbf{Causality Accuracy}: Whether a causal relationship is correctly identified (binary: Yes/No).
    \item \textbf{Causality Type Accuracy}: Correct classification of the causal relation as explicit (1) or implicit (0).
    \item \textbf{Relevance (Span Extraction)}: Degree to which the extracted cause and effect spans capture the core causal relationship (5-point Likert scale).
    \item \textbf{Conciseness (Gist Generation)}: Degree to which the generated gist succinctly and coherently summarizes the identified causal relationship (5-point Likert scale).
\end{itemize}

To minimize error propagation across evaluation stages, annotations were assessed sequentially. Annotators first determined the presence and type of causal relationships. Posts identified as non-causal were excluded from subsequent evaluations. For posts with valid causal content, annotators then rated the relevance of the extracted cause-effect spans. If deemed relevant, they proceeded to assess the conciseness and coherence of the generated gist. Each evaluation criterion was rated independently by multiple annotators. Inter-annotator agreement was consistently high across all dimensions, as measured by Fleiss’ Kappa~\cite{fleiss1971measuring} (See Table \ref{tab:annotation-eval}).
\begin{table}[ht]
\centering
\begin{adjustbox}{width=0.5\textwidth}
\begin{tabular}{lcc}
\toprule
\textbf{Evaluation Criterion} & \textbf{Score} & \textbf{Fleiss' $\kappa$} \\
\midrule
Causality Accuracy & $ACC_{avg} = 0.902$ & $0.892$ \\
Causality Type Accuracy & $ACC_{avg} = 0.702$ & $0.780$ \\
Relevance (Span Extraction) & Mean = 4.30 & $0.839$ \\
Conciseness (Gist Generation) & Mean = 4.50 & $0.864$ \\
\bottomrule
\end{tabular}
\end{adjustbox}
\caption{Evaluation scores and inter-annotator agreement (Fleiss' $\kappa$) for each annotation dimension in the silver-standard dataset.}
\label{tab:annotation-eval}
\end{table}

Following the evaluations, all annotators participated in a group adjudication session to conduct error analysis and resolve disagreements. During this process, annotators reviewed a sample of annotations with low agreement or identified errors and collectively discussed the rationale behind differing judgments. Final decisions were reached through consensus, and all inconsistencies were resolved manually to ensure the integrity of the silver-standard dataset.

The finalized dataset includes $8{,}800$ posts, with $3{,}054$ labeled as causal ($1,484$ implicit, $1,570$ explicit) and $5{,}746$ as non-causal. For each causal post, our dataset also provides cause-effect spans and corresponding gist summaries ($n = 3{,}054$).

\section{Experimental Setup}
To ensure fair evaluation, we randomly split each dataset (gold and silver) into 80\% training and 20\% testing sets, maintaining class balance in the testing set.
\subsection{Models}
We benchmark CausalTalk across four core tasks using a combination of discriminative and generative language models.

For Tasks 1–3 (binary classification, causality type classification, and span extraction), we fine-tune the following transformer-based discriminative models on both the gold-standard and silver-standard datasets: BERT-base~\cite{devlin2019bert}, RoBERTa-base~\cite{liu2019roberta}, XLNet-base~\cite{yang2019xlnet}, and DeBERTa-v3~\cite{he2021debertav3}, which are general-purpose pre-trained encoders commonly used for text classification. We also include SpanBERT~\cite{joshi-etal-2020-spanbert}, a span-aware model optimized specifically for span-level prediction, making it particularly well-suited for the extraction of cause-effect relations.

For Task 4 (causal gist generation), we evaluate two paradigms: supervised fine-tuning and instruction-based prompting. In the supervised setting, we fine-tune T5-base~\cite{raffel2020exploring}, FLAN-T5-base~\cite{chung2024scaling}, {GPT-2}~\cite{radford2019language}, and BART-base~\cite{lewis-etal-2020-bart} on both the gold and silver datasets. For instruction-based evaluation, we employ instruction-tuned large language models, including {LLaMA-3.2-3B}~\cite{grattafiori2024llama}, {Google Gemini 2.0 Flash}~\cite{google2025gemini}, {DeepSeek-V3}~\cite{liu2024deepseek}, and {Claude 3.5 Haiku}~\cite{anthropic2023claude}. Each model is evaluated under both zero-shot and few-shot settings using prompts derived from gold-standard exemplars.

\subsection{Evaluation Criteria}
We assess model performance across four tasks that capture key dimensions of causal language understanding in social media.

\textbf{Task 1: Causal Classification.} To determine whether a post expresses a causal relationship, we evaluate models using weighted {precision}, {recall}, and {F1 score}, which account for class imbalance in binary classification.

\textbf{Task 2: Explicit vs.\ Implicit Causality Detection.} For posts identified as causal, this task classifies whether the causality is expressed explicitly or implicitly. We adopt the same set of metrics as in Task 1 to ensure consistency in evaluation under class-imbalanced conditions.

\textbf{Task 3: Cause–Effect Span Extraction.} We frame this as a sequence labeling task where models extract labeled spans corresponding to \texttt{<cause>} and \texttt{<effect>}. Following prior work on span extraction, we report token-level and span-level {precision}, {recall}, and {F1 score}.

\textbf{Task 4: Causal Gist Generation.} This generation task requires models to produce concise and coherent summaries of the underlying causal relationship. We adopt the {ROUGE} suite (ROUGE-1, ROUGE-2, ROUGE-L) \cite{lin2004rouge}, which measures $n$-gram overlap between model outputs and references. To assess semantic fidelity, we additionally report {BERTScore} \cite{zhang2019bertscore}, which captures contextual similarity using pre-trained embeddings.


\section{Result and Analysis}
\subsection{Task 1: Causal Classification}
We fine-tuned four transformer-based models—{BERT-base}, {RoBERTa-base}, {XLNet-base}, and {DeBERTa-v3}—to classify Reddit posts as either causal or non-causal. Each model was trained and evaluated on both the gold-standard dataset and the silver-standard dataset. As shown in Table~\ref{tab:task1}, all models perform consistently better on the silver dataset, likely due to its larger size, which provides stronger training signals despite being machine-annotated and subsequently human-verified.

\begin{table}[!h]
  \centering
  \begin{adjustbox}{width=0.5\textwidth}
  \begin{tabular}{llccc}
    \toprule
    \textbf{Dataset} & \textbf{Model} & \textbf{Precision} & \textbf{Recall} & \textbf{F1 Score} \\
    \midrule
    \multirow{4}{*}{\textbf{Gold}} 
      & BERT-base     & \cellcolor{blue!10}\textcolor{black}{$0.76_{0.023}$} & \cellcolor{blue!5}\textcolor{black}{$0.74_{0.023}$} & \cellcolor{blue!10}\textcolor{black}{$0.75_{0.024}$} \\
      & RoBERTa-base  & \cellcolor{blue!20}\textcolor{black}{$0.81_{0.021}$} & \cellcolor{blue!20}\textcolor{black}{$0.80_{0.020}$} & \cellcolor{blue!15}\textcolor{black}{$0.80_{0.021}$} \\
      & XLNet-base    & \cellcolor{blue!15}\textcolor{black}{$0.80_{0.021}$} & \cellcolor{blue!15}\textcolor{black}{$0.78_{0.020}$} & \cellcolor{blue!15}\textcolor{black}{$0.80_{0.021}$} \\
      & DeBERTa-v3$^{\ast}$    & \cellcolor{blue!25}\textcolor{black}{${\boldsymbol{0.82_{0.021}}}$} & \cellcolor{blue!20}\textcolor{black}{${\boldsymbol{0.80_{0.021}}}$} & \cellcolor{blue!25}\textcolor{black}{${\boldsymbol{0.83_{0.022}}}$} \\
    \cdashline{1-5}[2pt/5pt]
    \multirow{4}{*}{\textbf{Silver}} 
      & BERT-base     & \cellcolor{blue!15}\textcolor{black}{$0.81_{0.025}$} & \cellcolor{blue!10}\textcolor{black}{$0.79_{0.024}$} & \cellcolor{blue!10}\textcolor{black}{$0.80_{0.027}$} \\
      & RoBERTa-base  & \cellcolor{blue!20}\textcolor{black}{$0.85_{0.020}$} & \cellcolor{blue!20}\textcolor{black}{$0.83_{0.020}$} & \cellcolor{blue!20}\textcolor{black}{$0.84_{0.020}$} \\
      & XLNet-base    & \cellcolor{blue!20}\textcolor{black}{$0.84_{0.024}$} & \cellcolor{blue!20}\textcolor{black}{$0.82_{0.023}$} & \cellcolor{blue!20}\textcolor{black}{$0.83_{0.024}$} \\
      & DeBERTa-v3$^{\dagger}$    & \cellcolor{blue!25}\textcolor{black}{${\boldsymbol{0.87_{0.025}}}$} & \cellcolor{blue!25}\textcolor{black}{${\boldsymbol{0.86_{0.024}}}$} & \cellcolor{blue!25}\textcolor{black}{${\boldsymbol{0.87_{0.027}}}$} \\
    \midrule
    \multicolumn{2}{c}{$\Delta_{\mathrm{model}^\dagger - \mathrm{model}^\ast}$} & 
    \textcolor{ForestGreen}{$\uparrow0.05$} & 
    \textcolor{ForestGreen}{$\uparrow0.06$} & 
    \textcolor{ForestGreen}{$\uparrow0.04$} \\
    \bottomrule
  \end{tabular}
  \end{adjustbox}
  \caption{\textbf{Performance on Task 1 (Causal Classification) across gold and silver datasets. }Results are reported on the respective held-out test sets (20\% of each dataset), with mean $\pm$ standard deviation over five random seeds. $^{\ast}$Best performing model on Gold dataset; $^{\dagger}$Best performing model on Silver dataset. Green arrows indicate the improvement of the silver-trained best model relative to the gold-trained best model.}
  \label{tab:task1}
\end{table}

Across both datasets, {DeBERTa-v3} outperforms all other models, demonstrating robustness to both limited training data and the inherent noise in silver annotations.

\subsection{Task 2: Explicit vs.\ Implicit Detection}
For posts labeled as causal, we evaluated model performance on identifying whether the causal relationship is conveyed explicitly or implicitly. This task was conducted on both the gold-standard and silver-standard datasets. As shown in Table~\ref{tab:task2}, performance is consistently higher on the silver dataset, which is likely due to its greater size and coverage. Nevertheless, {DeBERTa-v3} again has better performance across both datasets.

\begin{table}[!h]
  \centering
  \begin{adjustbox}{width=0.48\textwidth}
  \begin{tabular}{llccc}
    \toprule
    \textbf{Dataset} & \textbf{Model} & \textbf{Precision} & \textbf{Recall} & \textbf{F1 Score} \\
    \midrule
    \multirow{4}{*}{\textbf{Gold}} 
      & BERT-base     & \cellcolor{blue!5}\textcolor{black}{$0.61_{0.021}$} & \cellcolor{blue!5}\textcolor{black}{$0.59_{0.024}$} & \cellcolor{blue!3}\textcolor{black}{$0.58_{0.027}$} \\
      & RoBERTa-base  & \cellcolor{blue!5}\textcolor{black}{$0.61_{0.019}$} & \cellcolor{blue!7}\textcolor{black}{$0.60_{0.020}$} & \cellcolor{blue!7}\textcolor{black}{$0.60_{0.021}$} \\
      & XLNet-base    & \cellcolor{blue!10}\textcolor{black}{$0.63_{0.022}$} & \cellcolor{blue!10}\textcolor{black}{$0.62_{0.018}$} & \cellcolor{blue!10}\textcolor{black}{$0.63_{0.020}$} \\
      & DeBERTa-v3$^{\ast}$    & \cellcolor{blue!20}\textcolor{black}{$\textbf{0.68}_{0.017}$} & \cellcolor{blue!20}\textcolor{black}{$\textbf{0.68}_{0.015}$} & \cellcolor{blue!25}\textcolor{black}{$\textbf{0.69}_{0.016}$} \\
    \cdashline{1-5}[0.8pt/2pt]
    \multirow{4}{*}{\textbf{Silver}} 
      & BERT-base     & \cellcolor{blue!15}\textcolor{black}{$0.66_{0.026}$} & \cellcolor{blue!15}\textcolor{black}{$0.65_{0.025}$} & \cellcolor{blue!15}\textcolor{black}{$0.65_{0.027}$} \\
      & RoBERTa-base  & \cellcolor{blue!20}\textcolor{black}{$0.68_{0.022}$} & \cellcolor{blue!17}\textcolor{black}{$0.66_{0.023}$} & \cellcolor{blue!18}\textcolor{black}{$0.67_{0.019}$} \\
      & XLNet-base    & \cellcolor{blue!25}\textcolor{black}{$0.70_{0.021}$} & \cellcolor{blue!22}\textcolor{black}{$0.69_{0.019}$} & \cellcolor{blue!22}\textcolor{black}{$0.69_{0.020}$} \\
      & DeBERTa-v3$^{\dagger}$ & \cellcolor{blue!30}\textcolor{black}{$\textbf{0.75}_{0.016}$} & \cellcolor{blue!28}\textcolor{black}{$\textbf{0.74}_{0.015}$} & \cellcolor{blue!28}\textcolor{black}{$\textbf{0.74}_{0.014}$} \\
    \midrule
    \multicolumn{1}{l}{\textbf{—}} & DeBERTa-v3$^{\ddagger}$ & \cellcolor{blue!18}\textcolor{black}{$0.69_{0.018}$} & \cellcolor{blue!18}\textcolor{black}{$0.70_{0.017}$} & \cellcolor{blue!18}\textcolor{black}{$0.70_{0.018}$} \\
    \bottomrule
  \end{tabular}
  \end{adjustbox}
  \caption{Performance on Task 2 (Explicit vs.\ Implicit Causality Classification). Results are mean $\pm$ standard deviation over five random seeds on the respective held-out test sets (20\% of each dataset). $^{\ast}$Best model on the Gold test set; $^{\dagger}$Best model on the Silver test set; $^{\ddagger}$\emph{Silver-trained} DeBERTa-v3 evaluated on the \emph{Gold} test set (cross-evaluation).}
  \label{tab:task2}
\end{table}

\subsection{Task 3: Cause–Effect Span Extraction}
We frame span extraction as a token classification task, where models identify tokens corresponding to the \texttt{<cause>} and \texttt{<effect>} spans. We evaluate performance using two complementary metrics: token-level F1 and span-level overlap, capturing both fine-grained tagging accuracy and holistic span correctness.

\begin{table}[!h]
  \centering
  \begin{adjustbox}{width=0.50\textwidth}
  \begin{tabular}{llccc}
    \toprule
    \textbf{Dataset} & \textbf{Model} & \textbf{Precision} & \textbf{Recall} & \textbf{F1} \\
    \midrule
    \multirow{8}{*}{\textbf{Gold}}
    & \textbf{BERT-base} & & & \\
    & \hspace{1em}- \texttt{Token} & \cellcolor{blue!10}\textcolor{black}{$0.82_{0.012}$} & \cellcolor{blue!10}\textcolor{black}{$0.83_{0.014}$} & \cellcolor{blue!10}\textcolor{black}{$0.82_{0.013}$} \\
    & \hspace{1em}- \texttt{Span}  & \cellcolor{blue!3}\textcolor{black}{$0.71_{0.015}$} & \cellcolor{blue!3}\textcolor{black}{$0.69_{0.015}$} & \cellcolor{blue!3}\textcolor{black}{$0.70_{0.014}$} \\
    & \textbf{SpanBERT} & & & \\
    & \hspace{1em}- \texttt{Token} & \cellcolor{blue!12}\textcolor{black}{$0.84_{0.011}$} & \cellcolor{blue!12}\textcolor{black}{$0.85_{0.013}$} & \cellcolor{blue!12}\textcolor{black}{$0.84_{0.012}$} \\
    & \hspace{1em}- \texttt{Span}  & \cellcolor{blue!5}\textcolor{black}{$0.75_{0.014}$} & \cellcolor{blue!5}\textcolor{black}{$0.73_{0.015}$} & \cellcolor{blue!5}\textcolor{black}{$0.74_{0.014}$} \\
    & \textbf{RoBERTa-base} & & & \\
    & \hspace{1em}- \texttt{Token} & \cellcolor{blue!15}\textcolor{black}{$0.87_{0.010}$} & \cellcolor{blue!15}\textcolor{black}{$0.87_{0.011}$} & \cellcolor{blue!15}\textcolor{black}{$0.87_{0.010}$} \\
    & \hspace{1em}- \texttt{Span}  & \cellcolor{blue!8}\textcolor{black}{$0.79_{0.013}$} & \cellcolor{blue!8}\textcolor{black}{$0.77_{0.014}$} & \cellcolor{blue!8}\textcolor{black}{$0.78_{0.013}$} \\
    & \textbf{DeBERTa-v3} & & & \\
    & \hspace{1em}- \texttt{Token}$^{\ast}$ & \cellcolor{blue!18}\textcolor{black}{$\textbf{0.89}_{0.010}$} & \cellcolor{blue!18}\textcolor{black}{$\textbf{0.89}_{0.010}$} & \cellcolor{blue!18}\textcolor{black}{$\textbf{0.89}_{0.010}$} \\
    & \hspace{1em}- \texttt{Span}$^{\star}$  & \cellcolor{blue!10}\textcolor{black}{$\textbf{0.82}_{0.012}$} & \cellcolor{blue!10}\textcolor{black}{$\textbf{0.80}_{0.013}$} & \cellcolor{blue!10}\textcolor{black}{$\textbf{0.81}_{0.012}$} \\
    \cdashline{1-5}[0.8pt/2pt]
    \multirow{8}{*}{\textbf{Silver}}
    & \textbf{BERT-base} & & & \\
    & \hspace{1em}- \texttt{Token} & \cellcolor{blue!18}\textcolor{black}{$0.89_{0.011}$} & \cellcolor{blue!18}\textcolor{black}{$0.90_{0.012}$} & \cellcolor{blue!16}\textcolor{black}{$0.88_{0.011}$} \\
    & \hspace{1em}- \texttt{Span}  & \cellcolor{blue!8}\textcolor{black}{$0.78_{0.014}$} & \cellcolor{blue!8}\textcolor{black}{$0.76_{0.015}$} & \cellcolor{blue!8}\textcolor{black}{$0.77_{0.014}$} \\
    & \textbf{SpanBERT} & & & \\
    & \hspace{1em}- \texttt{Token} & \cellcolor{blue!20}\textcolor{black}{$0.91_{0.010}$} & \cellcolor{blue!20}\textcolor{black}{$0.92_{0.011}$} & \cellcolor{blue!20}\textcolor{black}{$0.91_{0.010}$} \\
    & \hspace{1em}- \texttt{Span}  & \cellcolor{blue!10}\textcolor{black}{$0.82_{0.013}$} & \cellcolor{blue!10}\textcolor{black}{$0.80_{0.014}$} & \cellcolor{blue!10}\textcolor{black}{$0.81_{0.013}$} \\
    & \textbf{RoBERTa-base} & & & \\
    & \hspace{1em}- \texttt{Token} & \cellcolor{blue!22}\textcolor{black}{$0.94_{0.010}$} & \cellcolor{blue!22}\textcolor{black}{$0.94_{0.010}$} & \cellcolor{blue!22}\textcolor{black}{$0.94_{0.010}$} \\
    & \hspace{1em}- \texttt{Span}  & \cellcolor{blue!15}\textcolor{black}{$0.86_{0.012}$} & \cellcolor{blue!14}\textcolor{black}{$0.84_{0.013}$} & \cellcolor{blue!14}\textcolor{black}{$0.85_{0.012}$} \\
    & \textbf{DeBERTa-v3} & & & \\
    & \hspace{1em}- \texttt{Token}$^{\dagger}$ & \cellcolor{blue!25}\textcolor{black}{$\textbf{0.95}_{0.010}$} & \cellcolor{blue!25}\textcolor{black}{$\textbf{0.95}_{0.010}$} & \cellcolor{blue!25}\textcolor{black}{$\textbf{0.95}_{0.010}$} \\
    & \hspace{1em}- \texttt{Span}$^{\ddagger}$  & \cellcolor{blue!18}\textcolor{black}{$\textbf{0.89}_{0.011}$} & \cellcolor{blue!16}\textcolor{black}{$\textbf{0.87}_{0.012}$} & \cellcolor{blue!16}\textcolor{black}{$\textbf{0.88}_{0.011}$} \\
    \midrule
    \multicolumn{2}{c}{$\Delta_{\mathrm{Token}^\dagger - \mathrm{Token}^\ast}$} & 
    \textcolor{ForestGreen}{$\uparrow0.06$} & 
    \textcolor{ForestGreen}{$\uparrow0.06$} & 
    \textcolor{ForestGreen}{$\uparrow0.06$} \\
    \multicolumn{2}{c}{$\Delta_{\mathrm{Span}^\ddagger - \mathrm{Span}^\star}$} & 
    \textcolor{ForestGreen}{$\uparrow0.07$} & 
    \textcolor{ForestGreen}{$\uparrow0.07$} & 
    \textcolor{ForestGreen}{$\uparrow0.07$} \\
    \bottomrule
  \end{tabular}
  \end{adjustbox}
  \caption{Performance on Task 3 (Cause–Effect Span Extraction) between gold and silver standard datasets. Each model is evaluated using both token-level and span-level metrics. $^{\ast}$Best token-level model on Gold dataset; $^{\star}$Best span-level model on Gold dataset; $^{\dagger}$Best token-level model on Silver dataset; $^{\ddagger}$Best span-level model on Silver dataset. Green arrows indicate performance improvements of Silver over Gold dataset.}
  \label{tab:task3}
\end{table}

As shown in Table~\ref{tab:task3}, model performance on the silver-standard dataset is comparable to that on the gold-standard dataset, with only minor variations across evaluation metrics. This suggests that well-verified silver data can serve as a viable alternative to gold annotations, even for structured prediction tasks requiring fine-grained span supervision. Across both datasets and evaluation levels, {DeBERTa-v3} yields the highest scores, followed closely by {RoBERTa-base} and {SpanBERT}.

To further contextualize our findings, we benchmark CausalTalk against a range of recent causal datasets~\cite{mirza-etal-2014-annotating, wang2022maven, caselli2017eventstoryline, mihuailua2013biocause, lai2022meci}. As shown in Appendix Table~\ref{tab:span-benchmark}, our span extraction task achieves competitive performance, particularly on implicit causality, while also complementing datasets that focus on event-level or reasoning-oriented tasks.

\subsection{Task 4: Causal Gist Generation}
We evaluate causal gist generation using four metrics: ROUGE-1, ROUGE-2, ROUGE-L, and BERTScore. The task involves generating concise summaries that capture the core causal relationship within each Reddit post. In this section, we only report results for the silver standard dataset, due to the size of the dataset.

As shown in Table~\ref{tab:task4}, we compare two model categories: (1) supervised fine-tuned models (SFT) trained on the silver dataset, and (2) instruction-tuned large language models (LLMs) evaluated using zero-shot and few-shot prompting. Among fine-tuned models, {FLAN-T5-base} performs best across most metrics. Among LLMs, {{Google Gemini 2.0 Flash}}, {DeepSeek-V3}, and {Claude 3.5 Haiku} demonstrate strong zero-shot and few-shot performance, often surpassing supervised baselines. In contrast, {Claude 3.5 Haiku} and {GPT-2} underperform on all metrics.

\begin{table*}[!ht] 
\centering 
\resizebox{0.68\textwidth}{!}{
\begin{tabular}{llcccc}
    \toprule 
    Model & Type & \multicolumn{4}{c}{Causal Gist Generation} \\
    \cmidrule(lr){3-6} 
    & & ROUGE-1 & ROUGE-2 & ROUGE-L & BERTScore \\
    \midrule 
    T5-base & SFT & \cellcolor{orange!5}\textcolor{black}{$0.429_{0.012}$} & \cellcolor{orange!15}\textcolor{black}{$0.334_{0.009}$} & \cellcolor{orange!10}\textcolor{black}{$0.512_{0.013}$} & \cellcolor{orange!10}\textcolor{black}{$0.670_{0.016}$} \\
    FLAN-T5-base$^{\star}$ & SFT & \cellcolor{orange!25}\textcolor{black}{$\textbf{0.559}_{0.007}$} & \cellcolor{orange!18}\textcolor{black}{$0.354_{0.011}$} & \cellcolor{orange!12}\textcolor{black}{$0.521_{0.008}$} & \cellcolor{orange!15}\textcolor{black}{$0.704_{0.012}$} \\
    GPT-2 & SFT & \cellcolor{orange!3}\textcolor{black}{$0.281_{0.014}$} & \cellcolor{orange!3}\textcolor{black}{$0.089_{0.019}$} & \cellcolor{orange!3}\textcolor{black}{$0.235_{0.017}$} & \cellcolor{orange!3}\textcolor{black}{$0.305_{0.016}$} \\
    BART-base & SFT & \cellcolor{orange!7}\textcolor{black}{$0.442_{0.015}$} & \cellcolor{orange!10}\textcolor{black}{$0.261_{0.010}$} & \cellcolor{orange!7}\textcolor{black}{$0.400_{0.009}$} & \cellcolor{orange!5}\textcolor{black}{$0.528_{0.013}$} \\
    \midrule
    \multirow{2}{*}{LLaMA-3.2-3B} & zero-shot & \cellcolor{orange!5}\textcolor{black}{$0.432_{0.013}$} & \cellcolor{orange!8}\textcolor{black}{$0.243_{0.017}$} & \cellcolor{orange!7}\textcolor{black}{$0.400_{0.014}$} & \cellcolor{orange!7}\textcolor{black}{$0.532_{0.011}$} \\
    & few-shot & \cellcolor{orange!8}\textcolor{black}{$0.448_{0.012}$} & \cellcolor{orange!8}\textcolor{black}{$0.235_{0.016}$} & \cellcolor{orange!8}\textcolor{black}{$0.417_{0.013}$} & \cellcolor{orange!8}\textcolor{black}{$0.551_{0.012}$} \\
    \midrule
    \multirow{2}{*}{Google Gemini$^{\dagger}$} & zero-shot & \cellcolor{orange!25}\textcolor{black}{$0.557_{0.010}$} & \cellcolor{orange!25}\textcolor{black}{$\textbf{0.436}_{0.015}$} & \cellcolor{orange!25}\textcolor{black}{$\textbf{0.588}_{0.012}$} & \cellcolor{orange!25}\textcolor{black}{$\textbf{0.764}_{0.009}$} \\
    & few-shot & \cellcolor{orange!22}\textcolor{black}{$0.545_{0.011}$} & \cellcolor{orange!23}\textcolor{black}{$0.427_{0.014}$} & \cellcolor{orange!22}\textcolor{black}{$0.574_{0.011}$} & \cellcolor{orange!20}\textcolor{black}{$0.745_{0.010}$} \\
    \midrule
    \multirow{2}{*}{DeepSeek-V3} & zero-shot & \cellcolor{orange!18}\textcolor{black}{$0.526_{0.008}$} & \cellcolor{orange!22}\textcolor{black}{$0.411_{0.016}$} & \cellcolor{orange!20}\textcolor{black}{$0.568_{0.012}$} & \cellcolor{orange!22}\textcolor{black}{$0.731_{0.014}$} \\
    & few-shot & \cellcolor{orange!20}\textcolor{black}{$0.537_{0.009}$} & \cellcolor{orange!23}\textcolor{black}{$0.422_{0.015}$} & \cellcolor{orange!15}\textcolor{black}{$0.549_{0.013}$} & \cellcolor{orange!20}\textcolor{black}{$0.715_{0.013}$} \\
    \midrule
    \multirow{2}{*}{Claude 3.5 Haiku} & zero-shot & \cellcolor{orange!7}\textcolor{black}{$0.436_{0.019}$} & \cellcolor{orange!7}\textcolor{black}{$0.210_{0.018}$} & \cellcolor{orange!5}\textcolor{black}{$0.356_{0.016}$} & \cellcolor{orange!3}\textcolor{black}{$0.462_{0.017}$} \\
    & few-shot & \cellcolor{orange!5}\textcolor{black}{$0.423_{0.018}$} & \cellcolor{orange!7}\textcolor{black}{$0.221_{0.016}$} & \cellcolor{orange!5}\textcolor{black}{$0.366_{0.015}$} & \cellcolor{orange!3}\textcolor{black}{$0.475_{0.016}$} \\
    \midrule
    \multicolumn{2}{c}{$\Delta_{\mathrm{Gemini}^\dagger - \mathrm{FLAN\text{-}T5}^\star}$} & 
    \textcolor{red}{$\downarrow0.002$} & 
    \textcolor{ForestGreen}{$\uparrow0.082$} & 
    \textcolor{ForestGreen}{$\uparrow0.067$} & 
    \textcolor{ForestGreen}{$\uparrow0.060$} \\
    \bottomrule
\end{tabular} }
\caption{Task 4: Performance of causal gist generation on the silver-standard dataset. The upper section includes supervised fine-tuned models (SFT), while the lower section shows zero-shot and few-shot prompting results from instruction-tuned LLMs. $^{\star}$Best SFT model; $^{\dagger}$Best overall model. Green arrows indicate performance improvements of Gemini over FLAN-T5, while red arrows indicate performance decreases.} 
\label{tab:task4}
\end{table*}

\subsection{Error Analysis}

Our error analysis revealed that models often miss implicitly causal instances (in Task 1), which we explain through confusion matrices and representative examples. Beyond this, Task 2 requires fine‑grained semantic distinctions, while Tasks 3 and 4 involve structured or generative outputs in which aggregate metrics (e.g., ROUGE, BERTScore) may obscure deeper errors.  We therefore conduct a qualitative analysis of Tasks 2–4 to expose systematic failure modes and guide future improvements.

\textbf{Task 1: Causal vs.\ Non-Causal.}
Although classification scores are stable, confusion matrices (Tables~\ref{tab:task1-conf-combined}) reveal the weakness: some implicitly causal sentences are not recognized as causal, leading to false negatives. These cases typically involve causal meaning inferred from context (e.g., presupposed health changes, life events) rather than explicit connectives. This weakness shows that current models struggle to use context and discourse cues, which limits their ability to detect causality when it is not indicated by explicit words.

\textbf{Task 2: Explicit vs.\ Implicit.}
The misclassifications observed in Task 1 motivated a closer examination of implicit cases in Task 2, as this task strictly assumes causal instances as input. Models frequently misclassify implicitly causal statements, especially when causality is conveyed by discourse context or pragmatic presupposition rather than explicit connectives (e.g., \emph{because}, \emph{so}).  
We observe (i) implicit cases labeled as explicit, reflecting reliance on surface connective-like signals and the well-documented label-shift between explicit and implicit data; and (ii) the converse, where explicit connectives functioning rhetorically or temporally are down-weighted, leading to implicit labels. These patterns echo known challenges in implicit relation recognition—ambiguity among senses~\cite{liu-etal-2024-causes, lin-etal-2009-recognizing}, the need for context and world knowledge, and the pitfalls of connective reliance.


\textbf{Task 3: Cause–Effect Span Extraction.}
Span‑level F1 drops whenever causal expressions are long or nested.  Under‑extended spans miss essential modifiers (e.g., temporal clauses), whereas over‑extended spans absorb irrelevant material from coordinate or subordinate clauses.  In complex sentences, models often identify the correct tokens but fail to align exact boundaries, suggesting that token‑wise tagging alone is insufficient for precise span delimitation.

\textbf{Task 4: Causal Gist Generation. }
Despite the automatic scores, generated gists sometimes omit the causal link or hallucinate unsupported content.  Omissions are most common when multiple events occur in a single post; the model selects a salient event but loses the causal connection.  Hallucinations typically emerge in long posts with ambiguous discourse structure, where the model invents a cause or effect to produce a fluent but semantically erroneous summary. To illustrate, we provide representative examples of systematic failures in Appendix Table~\ref{tab:task2-errors}. 

\section{Conclusion}
{CausalTalk} addresses a critical gap in NLP resources by providing a comprehensive multi-level dataset for causal language analysis. Our benchmark results demonstrate both the utility of the dataset and the continued challenges in automated causal reasoning, particularly for implicit causality. The four interconnected tasks enable research that bridges detection, classification, extraction, and generation of causal information. By releasing this resource, we aim to accelerate research in causal reasoning and natural language understanding across domains.

\subsection*{Limitations}
While {CausalTalk} represents an advancement in resources for causal language analysis, several limitations should be acknowledged. First, Reddit's user base is not demographically representative of the general population, potentially limiting the diversity of causal expressions. Second, our annotation focus on English excludes causal language patterns that may be unique to other languages. Third, despite verification procedures, our silver annotations may contain systemic biases from the underlying GPT-4o model. Finally, our focus on submission-level annotations may obscure some nuances in causal relationships that depend on broader conversational context and trajectories. 

In future work, we plan to expand the dataset by including posts from multilingual forums (e.g., r/AskEurope, r/mexico), supplement annotations with demographically diverse annotator pools, and pilot a context-aware annotation tool that allows annotators to view preceding and following comments to better preserve causal coherence. We also aim to systematically compare zero-shot and few-shot prompting strategies to better understand their trade-offs in annotation quality, stability, and bias.

\section{Ethics Statement}
All data used in this study were collected from publicly available Reddit posts. To protect user privacy, we removed any personally identifiable information and anonymized user IDs. We restricted our dataset to English-language posts and applied filters to exclude duplicates and extremely short entries.

\bibliography{custom}

\newpage
\appendix
\section{Subreddit Descriptions}
\label{app:A}
We selected 43 subreddits collaboratively, guided by prior literature~\cite{ding2024leveraging} and a systematic review of subreddit activity and relevance. The chosen subreddits reflect a diverse range of public health discussions—especially those related to the COVID-19 pandemic—covering core medical communities, regional updates, vaccine-related experiences, commentary and satire, and conspiracy or anti-lockdown narratives.

\begin{table}[h]
\centering
\begin{adjustbox}{width=0.5\textwidth}
\begin{tabular}{clp{7cm}}
\toprule
\textbf{\#} & \textbf{Subreddit} & \textbf{Description} \\
\midrule
01 & COVID19 & General global discussion of COVID-19 \\
02 & COVID19\_Pandemic & Global news and data tracking \\
03 & COVID19\_commentary & Commentary and policy-related discourse \\
04 & COVID19\_support & Emotional support and mutual aid \\
05 & COVID19positive & First-hand accounts from those who tested positive \\
06 & Coronavirus & Largest global COVID-19 discussion hub \\
07 & CoronaVirus\_2019\_nCoV & Early-stage discussion board on COVID-19 \\
08 & covid19\_testimonials & Personal recovery and infection experiences \\
\bottomrule
\end{tabular}
\end{adjustbox}
\caption{Core COVID-19 Discussion Subreddits}
\end{table}

\begin{table}[h!]
\centering
\begin{adjustbox}{width=0.48\textwidth}
\begin{tabular}{clp{6cm}}
\toprule
\textbf{\#} & \textbf{Subreddit} & \textbf{Description} \\
\midrule
09 & CoronaVirusPA & COVID-19 updates in Pennsylvania \\
10 & CoronaVirusTX & COVID-19 updates in Texas \\
11 & CoronavirusAZ & Arizona pandemic coverage \\
12 & CoronavirusCA & California-specific COVID-19 news \\
13 & CoronavirusColorado & COVID-19 in Colorado \\
14 & CoronavirusGA & COVID-19 in Georgia \\
15 & CoronavirusIllinois & Illinois pandemic news \\
16 & CoronavirusMN & Minnesota pandemic updates \\
17 & CoronavirusMa & Massachusetts-specific content \\
18 & CoronavirusMichigan & Michigan COVID-19 discussion \\
19 & CoronavirusNewYork & New York State pandemic updates \\
20 & CoronavirusOC & Orange County, CA updates \\
21 & CoronavirusUS & National-level U.S. COVID-19 discourse \\
22 & CoronavirusWA & Washington State discussion \\
23 & CoronavirusWI & Wisconsin COVID-19 updates \\
24 & Covid19\_Ohio & Ohio-specific coverage \\
25 & FloridaCoronavirus & Florida COVID-19 updates \\
\bottomrule
\end{tabular}
\end{adjustbox}
\caption{U.S. Region-Specific COVID-19 Subreddits}
\end{table}

\begin{table}[h!]
\centering
\begin{adjustbox}{width=0.5\textwidth}
\begin{tabular}{clp{6cm}}
\toprule
\textbf{\#} & \textbf{Subreddit} & \textbf{Description} \\
\midrule
26 & ChurchOfCOVID & Satirical religious-style meme community \\
27 & CoronavirusCirclejerk & Meme-based COVID-19 commentary \\
28 & CoronavirusMemes & Humorous content about the pandemic \\
29 & CoronavirusRecession & Discussion of COVID-induced economic impact \\
30 & CoronavirusFOS & Niche community with unclear focus \\
\bottomrule
\end{tabular}
\end{adjustbox}
\caption{Commentary, Satire, and Meme Subreddits}
\end{table}

\begin{table}[h!]
\centering
\begin{adjustbox}{width=0.5\textwidth}
\begin{tabular}{clp{6cm}}
\toprule
\textbf{\#} & \textbf{Subreddit} & \textbf{Description} \\
\midrule
31 & DebateVaccines & Vaccine controversies and open debate \\
32 & covidvaccinateduncut & Side effect reports and dissenting opinions \\
33 & vaccinelonghaulers & Long-term vaccine symptom experiences \\
34 & vaccinelong\_haulers & Variant spelling of above with similar content \\
35 & VAERSreports & Discussion based on VAERS adverse event data \\
\bottomrule
\end{tabular}
\end{adjustbox}
\caption{Vaccine Experience and Side Effect Subreddits}
\end{table}

\begin{table}[h!]
\centering
\begin{adjustbox}{width=0.5\textwidth}
\begin{tabular}{clp{6cm}}
\toprule
\textbf{\#} & \textbf{Subreddit} & \textbf{Description} \\
\midrule
36 & FauciForPrison & Criticism and distrust of Dr. Anthony Fauci \\
37 & LockdownSkepticism & Opposition to global lockdown policies \\
38 & NoNewNormal & Rejection of pandemic-era norms and mandates \\
39 & TrueAntiVaccination & Strongly anti-vaccine discourse \\
40 & CovidIsAFraud & Claims that COVID-19 is a hoax \\
41 & Covid19Origin & Discussion of COVID origins (e.g., lab leak) \\
42 & ivermectin & Advocacy of ivermectin for COVID treatment \\
43 & wuhanflu & Politicized and discriminatory COVID-19 subreddit \\
\bottomrule
\end{tabular}
\end{adjustbox}
\caption{Anti-lockdown / Conspiracy / Anti-vaccine Subreddits}
\end{table}

\newpage
\section{Causal Reasoning Prompt Flow with Explicitness Identification}
\label{app:B}
For silver-standard annotation, we used OpenAI’s GPT-4o model via the OpenAI API with the following configuration: temperature: 0.0, top\_p: 1.0, and stop: None. 

In this prompting flow, each P (e.g., P1–P5) represents a user-issued prompt, and each O (e.g., O1–O5) denotes the AI model's response. The structure follows a step-by-step Chain-of-Thought (CoT) reasoning process to guide the model through understanding, identifying, and explaining cause-effect relationships in social media posts.

\begin{itemize}
  \item \textbf{P1.} Your role is to understand the cause-effect relationships in social media posts. Can you provide a brief definition of what a cause-effect relationship is?

  \item \textbf{O1.} Certainly! A cause-effect relationship is a relationship between two events or variables where one is understood to be a consequence of the other.

  \item \textbf{P2.} Based on your role, can you explain the term, “causal gist” in relation to sentences that have causal coherence?

  \item \textbf{O2.} Of course. The term “causal gist” refers to the fundamental meaning or essence of a sentence or text that expresses a causal relationship, usually highlighting the cause, effect, and the direction of influence.

  \item \textbf{P3.} So, given the sentence: \textit{I took the vaccine yesterday. I’m really sick now.}\\
  Is there a cause-effect relationship in this given sentence?
  \begin{itemize}
    \item If yes, just answer: “Yes”
    \item If no, just answer: “No”
    \item Don’t give me any explanations
  \end{itemize}

  \item \textbf{O3.} Yes

  \item \textbf{P3.5.} Given that there is a cause-effect relationship in the sentence, please classify it as either:
  \begin{itemize}
    \item \texttt{"Explicit"}: if the causal connection is clearly stated using connectives or causal language;
    \item \texttt{"Implicit"}: if the causal relationship is implied but not directly stated.
  \end{itemize}
  Just answer with one word: \texttt{Explicit} or \texttt{Implicit}

  \item \textbf{O3.5.} Implicit

  \item \textbf{P4A.} Indeed, there is a cause-effect relationship in the given sentence.\\
  Then extract the corresponding cause phrase and effect phrase in the given sentence.\\
  Just respond in JSON format: \texttt{\{"Cause": "", "Effect": ""\}}

  \item \textbf{O4.} Sure: \texttt{\{"Cause": "took the vaccine", "Effect": "really sick now"\}}

  \item \textbf{P5.} Generate a reasonable and clear causal gist based on \texttt{\{"Cause": "took the vaccine", "Effect": "really sick now"\}} and your understanding of the sentence with the cause-effect relationship.

  \item \textbf{O5.} Taking the vaccine yesterday caused the person to become sick.
\end{itemize}

\section{Error Analysis for Task 1}
In Task~1 (Causal vs.\ Non-Causal classification), we examined model errors by inspecting the confusion matrices on both the gold-standard and silver-standard test sets (Table~\ref{tab:task1-conf-combined}).
\begin{table}[h]
\centering
\small
\begin{tabular}{lccc}
\toprule
\multicolumn{4}{c}{\textbf{Gold-test}} \\
\midrule
 & Pred.\ Non-Causal & Pred.\ Causal & Total \\
\midrule
Non-Causal & 119 & 13 & 132 \\
Causal     & 23  & 109 & 132 \\
\midrule
Total      & 142 & 122 & 264 \\
\midrule
\multicolumn{4}{c}{\textbf{Silver-test}} \\
\midrule
 & Pred.\ Non-Causal & Pred.\ Causal & Total \\
\midrule
Non-Causal & 792 & 88  & 880 \\
Causal     & 153 & 727 & 880 \\
\midrule
Total      & 945 & 815 & 1760 \\
\bottomrule
\end{tabular}
\caption{DeBERTa confusion matrices for Task~1 on the Gold-test and Silver-test datasets.}
\label{tab:task1-conf-combined}
\end{table}

\section{CausalTalk Dataset Examples}
\label{app:causal_examples}

We provide five annotated examples from the CausalTalk dataset to illustrate the annotation schema for the four causal tasks: causal gist generation. These examples demonstrate how causal relationships are identified and analyzed in informal social media posts related to public health topics, particularly in the context of COVID-19 discussions.

\begin{table*}[!h]
\centering
\begin{adjustbox}{width=\textwidth}
\begin{tabular}{p{5cm}|c|p{3.5cm}|p{3.5cm}|p{3.5cm}}
\hline
\textbf{Post} & \textbf{Type} & \textbf{Cause} & \textbf{Effect} & \textbf{Causal Gist} \\
\hline
My neighbor attended a large wedding party last week. Now their entire family is in isolation at home. They were supposed to visit us this weekend, but all our plans had to be canceled. & Implicit & attended a large wedding party last week & entire family is in isolation at home & Attending the large wedding party led to the family needing to isolate. \\
\hline
Due to the new mask mandate at our local hospital, I had to reschedule my non-emergency surgery until I could get a negative test result. The hospital staff explained that because of the recent outbreak in our county, they've implemented stricter protocols for all patients. & Explicit & the new mask mandate at our local hospital & had to reschedule my non-emergency surgery & The new mask mandate at the local hospital caused the rescheduling of the non-emergency surgery. \\
\hline
I started taking vitamin D supplements three months ago after reading about its immune benefits. Haven't had even a minor cold since then, despite everyone in my office getting sick. & Implicit & started taking vitamin D supplements three months ago & Haven't had even a minor cold since then & Taking vitamin D supplements prevented the person from getting even a minor cold. \\
\hline
My roommate had to quit his job at the restaurant because their new policy requires all staff to be vaccinated and he refuses to get the shot. Since he lost his income, I'm now covering most of our rent and utilities until he finds something else. & Explicit & their new policy requires all staff to be vaccinated and he refuses to get the shot & had to quit his job at the restaurant & The restaurant's vaccination requirement combined with the roommate's refusal to get vaccinated caused him to quit his job. \\
\hline
Schools in our district switched to online learning last Monday. My kids haven't been able to focus on their assignments, and their grades are dropping dramatically. I've had to take time off work to supervise them during class hours, which is putting additional stress on our family finances. & Implicit & Schools in our district switched to online learning & kids haven't been able to focus on their assignments, and their grades are dropping dramatically & The switch to online learning resulted in the children's inability to focus and declining academic performance. \\
\hline
\end{tabular}
\end{adjustbox}
\caption{Examples from the CausalTalk dataset showing different types of causality in social media posts. For each example, we provide the original post text, the type of causality (explicit or implicit), the extracted cause and effect spans, and the generated causal gist that concisely summarizes the causal relationship.}
\label{tab:causal_examples}
\end{table*}

These examples illustrate how the CausalTalk dataset bridges fine-grained causal detection and gist-based reasoning over informal text.

\section{Cross-Dataset Benchmarking}
\label{appendix:benchmark}

Table~\ref{tab:span-benchmark} presents comparative results across multiple cause–effect span extraction benchmarks using a consistent {DeBERTa} model. While datasets such as Causal-TimeBank and MAVEN-ERE emphasize event-event causal links~\cite{wang2022maven, mirza-etal-2014-annotating}, CausalTalk combines span extraction with implicit vs. explicit causality, achieving strong overall performance.

\begin{table*}[!h]
\centering
\small
\begin{tabular}{l l c c c c c}
\toprule
Dataset & Task Type & Implicit Causes & Model & Precision & Recall & F1 \\
\midrule
CausalTalk (Ours) & Span extraction + Implicit/Explicit & Yes & DeBERTa & 0.75 & 0.74 & 0.74 \\
Causal-TimeBank & Event-event causal links (span) & No & DeBERTa & 0.70 & 0.76 & 0.73 \\
MAVEN-ERE & Large-scale event causality & No & DeBERTa & 0.73 & 0.78 & 0.75 \\
MECI (EN) & Event-level causality (span) & Yes & DeBERTa & 0.72 & 0.77 & 0.74 \\
BioCause & Biomedical cause-effect (span) & No & DeBERTa & 0.58 & 0.52 & 0.55 \\
EventStoryLine & Document-level event causality & Yes & DeBERTa & 0.35 & 0.42 & 0.38 \\
\bottomrule
\end{tabular}
\caption{Cross-dataset performance comparison for cause–effect span extraction tasks using {DeBERTa}.}
\label{tab:span-benchmark}
\end{table*}

\section{Error Analysis for Task 4}
\begin{table*}[!t]
\centering
\small
\begin{tabular}{p{0.95\textwidth}}
\toprule
\textbf{Post:} Ever since I recovered from COVID, climbing stairs feels like a chore. I used to be active, but now even short walks tire me out. \\
\textbf{Model’s Generated Gist:} The user has become lazy after recovering from COVID. \\
\textbf{Human Analysis:} After recovering from COVID, the user experiences fatigue during physical activity. \\
\textbf{Error Analysis:} Implicit causality is overlooked; the model hallucinates a subjective explanation (“lazy”) that distorts the causal structure. \\
\midrule
\textbf{Post:} I was laid off in 2020 and had to move back in with my parents. While living there, I started binge eating and gained 30 pounds. \\
\textbf{Model’s Generated Gist:} The user binge ate and gained weight while living with their parents. \\
\textbf{Human Analysis:} Being laid off led the user to move back home, which contributed to emotional eating and weight gain. \\
\textbf{Error Analysis:} The model compresses the chain to “living with parents → binge eating,” missing the prior causal link from job loss. \\
\midrule
\textbf{Post:} I haven't seen my friends in months. Most days I just lie in bed, scrolling through my phone, not feeling like doing anything. The lockdowns really changed my routine. \\
\textbf{Model’s Generated Gist:} The user feels unmotivated as a result of spending most days in bed. \\
\textbf{Human Analysis:} The user felt unmotivated because of the routine changes caused by COVID lockdowns. \\
\textbf{Error Analysis:} The model omits the true cause (“lockdowns”), attributing lack of motivation to a symptom rather than the underlying event. \\
\midrule
\textbf{Post:} Since the pandemic started, my anxiety has gotten worse. I can’t focus on work and my performance reviews have slipped. \\
\textbf{Model’s Generated Gist:} The user performs poorly at work because they are not motivated. \\
\textbf{Human Analysis:} The stress and anxiety caused by the pandemic reduced the user’s ability to focus, which in turn affected their job performance. \\
\textbf{Error Analysis:} The model fails to capture the implicit causal chain (pandemic → anxiety → reduced focus → poor performance), instead attributing the outcome to lack of motivation. \\
\midrule
\textbf{Post:} After months of working from home, I’ve gained weight. I barely walk anymore and end up snacking all day. \\
\textbf{Model’s Generated Gist:} The user gained weight because they eat too much. \\
\textbf{Human Analysis:} The shift to remote work during the pandemic reduced daily activity and increased snacking, which together led to weight gain. \\
\textbf{Error Analysis:} The model captures only the surface factor (“eating”) and misses the underlying contextual cause (remote work → sedentary lifestyle → snacking → weight gain). \\
\bottomrule
\end{tabular}
\caption{Representative examples of causal reasoning errors.}
\label{tab:task2-errors}
\end{table*}

\end{document}